\pdfoutput=1
\documentclass[11pt]{article}
\usepackage[final]{acl}

\usepackage{times}
\usepackage{latexsym}
\usepackage{graphicx}
\usepackage{float}
\usepackage{color, colortbl}
\definecolor{Gray}{gray}{0.9}
\usepackage{graphicx}
\usepackage{makecell}
\usepackage{lipsum}
\usepackage{float}
\usepackage{booktabs}
\usepackage{placeins}
\usepackage{hyperref}
\usepackage{epstopdf}
\usepackage{comment}
\usepackage{amsmath}
\usepackage{bm} % For bold math symbols
\usepackage{algorithm2e}
\usepackage{amsmath}
\usepackage{pifont}
\usepackage{wasysym}
\usepackage{amssymb}
\usepackage{multirow}
\usepackage{subcaption}
\usepackage[T1]{fontenc}
\usepackage[utf8]{inputenc}

\usepackage{microtype}

\title{{Detecting Future-related Contexts of Entity Mentions}}
\author{
    Puneet Prashar \and Krishna Mohan Shukla \\
    Rajiv Gandhi Institute of Petroleum Technology \\
    \texttt{\{21mc3012, shuklakrishnamohan14\}@rgipt.ac.in} 
    \AND
    Adam Jatowt \\
    University of Innsbruck \\
    Innsbruck, Austria \\
    \texttt{adam.jatowt@uibk.ac.at}
}

\newcommand{\keywords}[1]{\textbf{Keywords:} #1} 
\begin{document}
\maketitle
The ability to automatically identify whether an entity is referenced in a future context can have multiple applications including decision making, planning and trend forecasting. 
This paper focuses on detecting implicit future references in entity-centric texts, addressing the growing need for automated temporal analysis in information processing. We first present a novel dataset of 19,540 sentences
%\footnote{The dataset will be publicly released after paper publication.}
built around popular entities sourced from Wikipedia, which consists of future-related and non-future-related contexts in which those entities appear. As a second contribution, we evaluate the performance of several Language Models including also Large Language Models (LLMs) on the task of distinguishing future-oriented content in the absence of explicit temporal references. 


\section{Introduction}
The rapid growth of digital information has made temporal analysis of textual content increasingly crucial for various applications from strategic planning to trend analysis. While traditional natural language processing approaches have explored variets of named entity oriented tasks such as named entity recognition, classification or linking \cite{keraghel2024survey}, the temporal aspect of entities, particularly their future orientation, remains still a relatively underexplored domain. This paper introduces an approach specifically designed to detect and classify future references in entity-centric texts.

The ability to automatically identify whether an entity is referenced in a future context has significant implications across multiple domains \cite{saha2025wisdomcrowdsforecastingforecast}. For instance, in corporate intelligence, understanding future plans related to companies can provide valuable insights for strategic decision-making. Similarly, in geographical analysis, detecting future developments associated with locations can aid urban planning and development strategies. However, this task presents unique challenges as future references often rely on subtle contextual cues rather than explicit markers \cite{nakajima2014investigation,nakajima2020future,DBLP:conf/wsdm/RegevJ024}. 

Traditional approaches to temporal text analysis and extraction have primarily focused on explicit time expressions \cite{saha2025wisdomcrowdsforecastingforecast,dias2011future,jatowt2011extracting}. These methods, while effective for historical and present-time analysis, often fall short when dealing with future references, which can be implicit and more nuanced. Furthermore, existing datasets and methodologies typically concentrate on general temporal relations rather than specifically targeting future references in entity-centric text. 

We addresses such limitations by providing lightweight and effective approaches to detecting and collecting future-related statements with entity mentions which can contain both implicit and explicit markers of future orientation. Such statements can be then later subject to different forms of aggregation and summarization to allow generating effective forecasts based on the wisdom of the crowd assumption \cite{saha2025wisdomcrowdsforecastingforecast}. Additionally, an approach for detecting future-related contexts can be helpful for building search engines and improving retrieval technologies for future-oriented content \cite{kanhabua2011ranking,baeza2005searching,kawai2010chronoseeker}.

We also note that although there have been several investigations done lately on probing whether LLMs can be effective in future forecasting \cite{schoenegger2023large,zhang2024largelanguagemodelsevent,mutschlechner2025analyzing,zou2022forecasting,nako2025navigatingtomorrowreliablyassessing,yuan2023back,jin2020forecastqa}, generative AI is still prone to hallucinations and its black box character makes it difficult to transparently produce effective and reliable forecasts.   

Our work addresses such limitations through the following contributions:
\begin{enumerate}

\item We present a novel dataset of carefully curated 19,540 sentences, balanced between future and non-future-related sentences, that was built around high-impact entities sourced from the Wikipedia. In our task setup, we mask any explicit date references, so that the model is forced to learn deeper contextual patterns that indicate future orientation. 
\item We establish comprehensive baselines using traditional machine learning approaches, including Decision Trees, Random Forest, Naive Bayes, and Support Vector Machines (SVM). We then conduct extensive experiments with transformer-based models including BERT, RoBERTa, DeBERTa, and ALBERT to distinguish future-related from non-future-related contexts of entities. 
\item We finally explore the capabilities of several Large Language Models (LLMs) including FLAN-T5, Llama 3 and Mistral in this temporal classification task.

\end{enumerate}

\section{Dataset Creation}

In this paper, we define future-related sentences as sentences that discuss events or states expected to occur after the time at which they were created (i.e., after their publication dates)\footnote{Note that it is not necessary to actually know the date of content creation or publication since humans can infer the time order from implicit signals in the content and from the overall context.}. 

\begin{figure*}[t!]
    \centering
    \includegraphics[width=.8\textwidth]{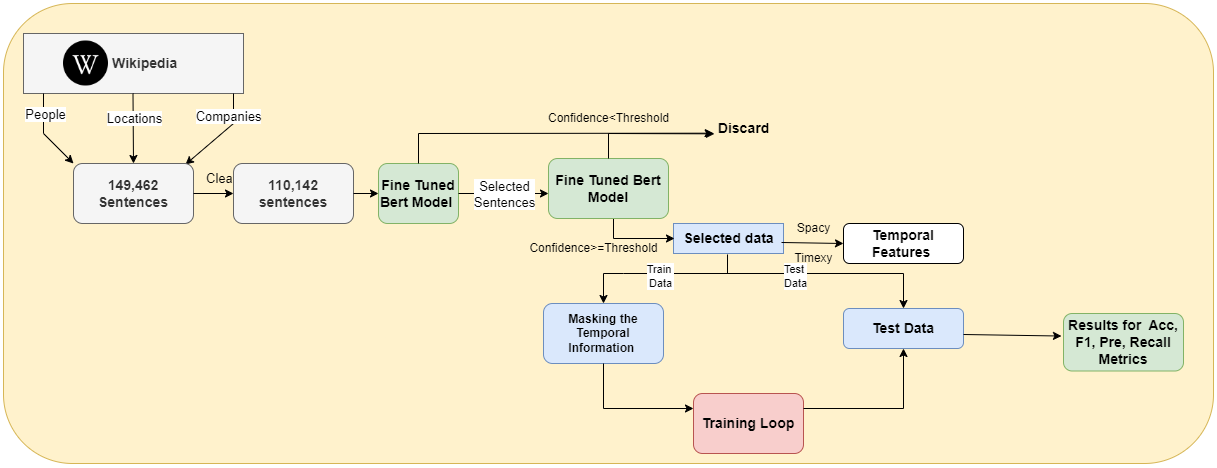}
    \caption{Pipeline of dataset creation and model training.}
    \label{measurement_setup}
\end{figure*}

Since, to the best of our knowledge, there was no ready, dedicated dataset for future-related information extraction about entities, we have decided to construct a dedicated dataset using the pipeline shown in Fig. \ref{measurement_setup}.
We focused on entities across three primary categories: notable individuals, geographical locations, and corporate/institution organizations. Our data collection process began with a careful selection of 300 distinct entities, comprising 100 entries from each category. For the people category, we selected prominent individuals spanning various domains including but not limited to political leaders, scientific innovators, artistic personalities, and business executives. The locations category incorporated a wide spectrum of geographical entities, including major metropolitan areas, countries, historical landmarks, and regions of significant cultural or economic importance. The last category consisted of organizations representing diverse sectors, varying in scale from multinational corporations to influential regional enterprises.

Based on the gathered entities, we collected 149,462 sentences from Wikipedia along with their previous sentence, next sentences and the URLs of the Wikipedia articles. To scrape these sentences we used Wikipedia API's search function. We found sentences related to entities by using their names as queries. We filtered out sentences containing noise by using regular expressions and hand-crafted rules\footnote{We removed sentences containing excessive symbols (e.g., >5 special characters), HTML/XML markup, URLs, random alphanumeric strings, and repeated characters, as well as sentences with less than 3 words or over 150 words.}.

We next classified sentences into future-related ones or not. The classification involved the application of BERT model \cite{DBLP:journals/corr/abs-1810-04805} fine-tuned on the Chronicle-2050 dataset \cite{DBLP:conf/wsdm/RegevJ024} which contains 6,800 manually curated and labeled sentences as either future-related or non-future-related. 

To ensure high quality, we applied a confidence threshold of 70\%. This resulted in 31,137 labeled sentences. We then again finetuned the same BERT model using the gathered 31,137 sentences and their labels. Next we run the updated BERT model again on the entire data with the same confidence threshold. After this step we obtained 9,540 sentences judged as future-related while  17,331 sentences were labeled as non future-related. We decided for the two-stage approach as manually annotating a large dataset would be quite costly. 

To validate the reliability of the automated classification system, we randomly selected 2,000 sentences equally distributed between future-related and non-future-related categories. Two annotators manually annotated these sentences. The classification model had a Cohen's Kappa score of 0.76 when compared to the human annotations, indicating a strong agreement between human and model judgments. The final dataset contains 19,540 sentences. This includes all 9,540 sentences that were classified as future-related with high confidence, complemented by a random selection of 10,000 sentences from the non-future-related category.

\begin{figure*}[t]
    \begin{minipage}[t]{0.4\textwidth}
        \centering
        \includegraphics[width=\textwidth]{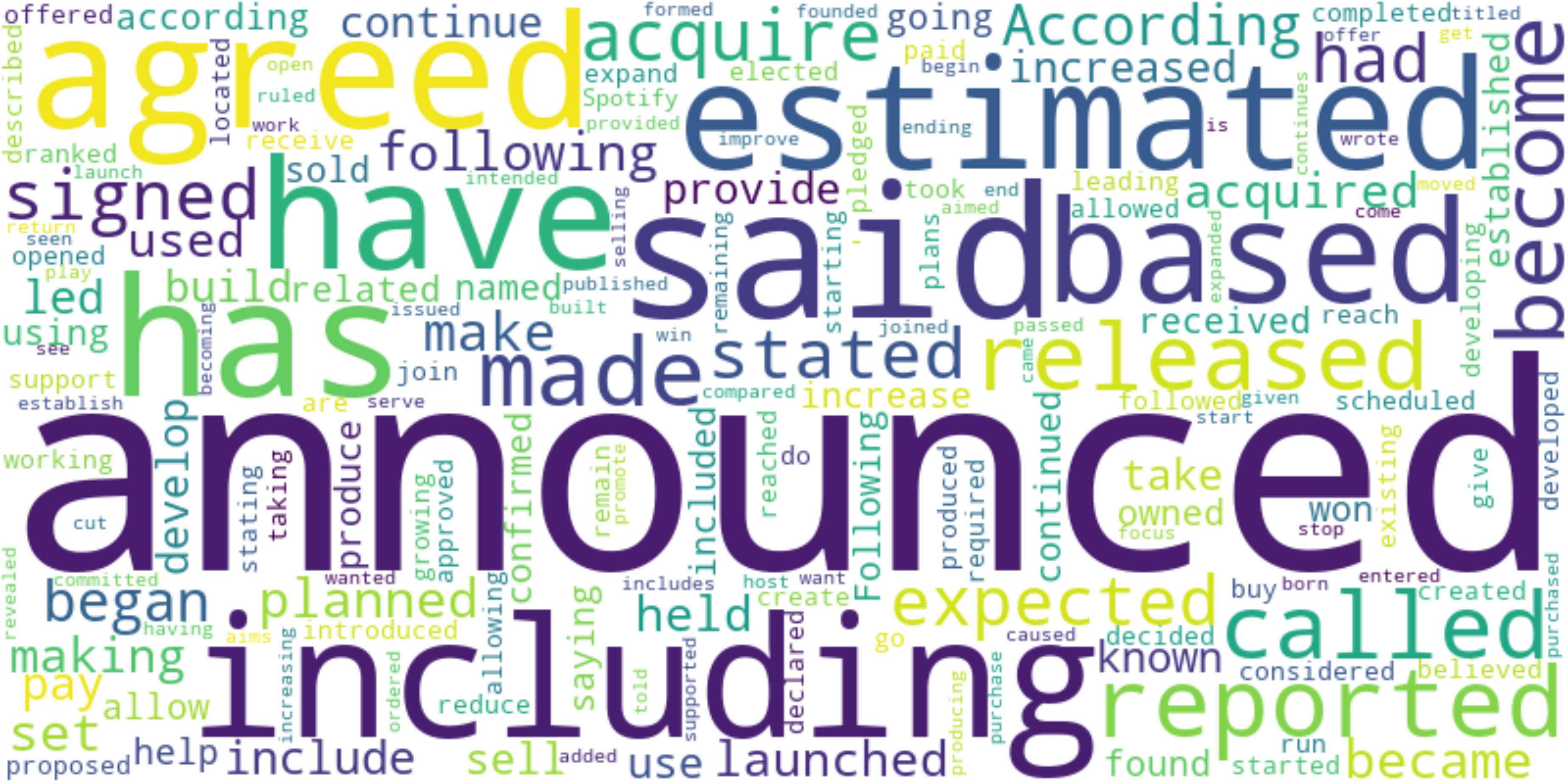}
        \caption{Top-verbs appearing in future-related sentences.}
        \label{verb_freq}
    \end{minipage}
    \hfill
    \begin{minipage}[t]{0.4\textwidth}
        \centering
        \includegraphics[width=\textwidth]{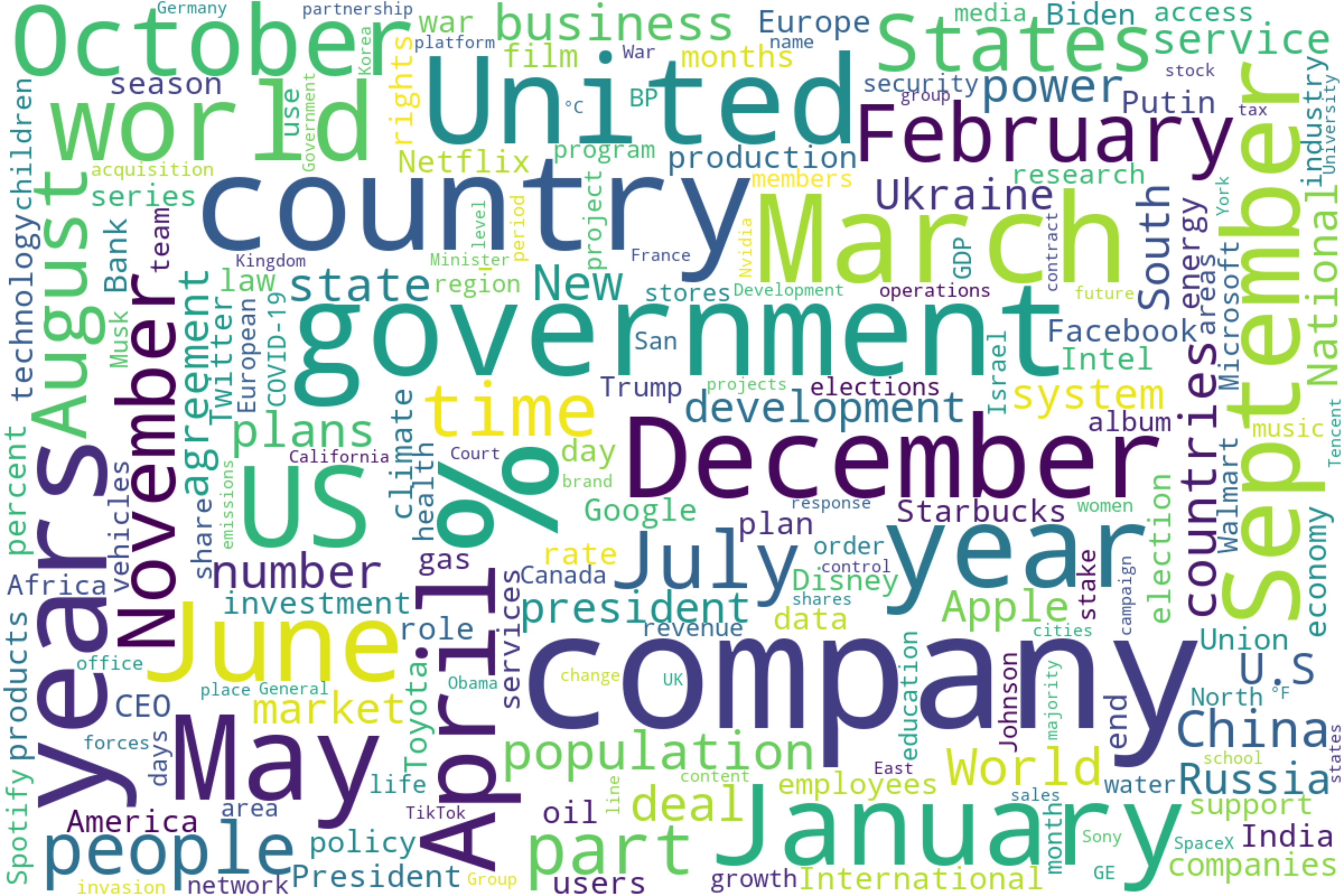}
        \caption{Top-nouns appearing in future-related sentences.}
        \label{noun_freq}
    \end{minipage}
\end{figure*}

We next employed the Timexy temporal expression detector\footnote{\url{https://pypi.org/project/timexy/}} to identify and categorize explicit temporal expressions within each sentence including both absolute dates and relative temporal expressions, as well as calendar references of varying specificity. We also used Spacy
%\footnote{https://spacy.io/} 
and NLTK
%\footnote{https://www.nltk.org/} 
to extract verbs, nouns, objects, subjects, and predicates. In Figure \ref{verb_freq} and Figure \ref{noun_freq} we show respectively, word clouds of verbs and nouns constructed from the future-related sentences.
%In the final dataset, location entites are in 8,983 sentences (3,810 (42.41\%) future sentences and 5,173 (57.59\%) non-future sentences).
%For the Person entity, the dataset contains 5,428 sentences (2,393 (44.09\%) future sentences and 3,035 (55.91\%) non-future sentences).
%Finally, the Company entity is in 5,129 sentences (3,337 (65.06\%) future sentences and 1,792 (34.94\%) non-future sentences).

\section{Method}

To preprocess the data for classification we replaced direct date mentions, year references, and numerical temporal indicators with generic temporal tokens (e.g., converting "2025" to "[DATE]"), also masking the contextual temporal information, such as phrases like "in the coming years," "plans to," or modal verbs indicating future intent. This is to ensure that tested models develop a deeper understanding of future orientation based on semantic and contextual clues rather than relying solely on explicit temporal markers.

The first category of models we test comprises traditional machine learning approaches, including Decision Trees, Random Forest, Naive Bayes, Support Vector Machines (SVM), and Logistic Regression. For these classical models, we extracted TF-IDF vectors from the unmasked text, syntactic features derived from dependency parsing, indicators of modal verb presence, and various part-of-speech patterns.
\begin{figure}[t]
    \centering
    \begin{minipage}[t]{0.4\textwidth}
        \centering
        \includegraphics[width=\textwidth]{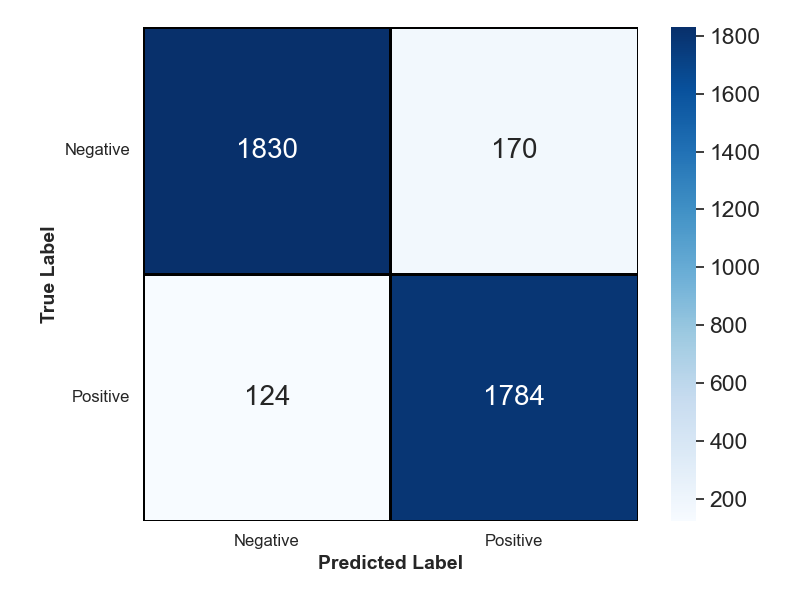}
        \caption{Confusion Matrix of RoBERTa classifier.}
        \label{confusion_matrix}
    \end{minipage}
    \hfill
    \begin{minipage}[t]{0.35\textwidth}
        \centering
        \includegraphics[width=\textwidth]{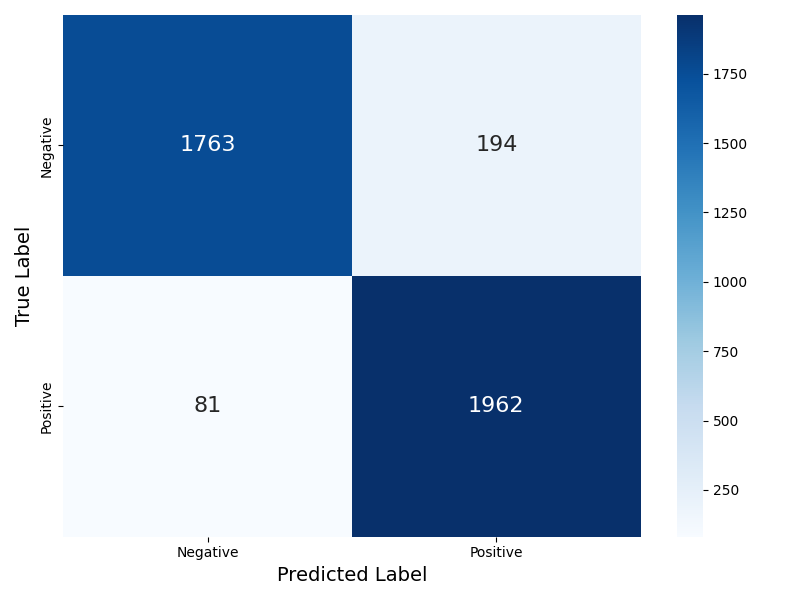}
        \caption{Confusion Matrix of Finetuned LLama3.}
        \label{confusion_matrix-llama3}
    \end{minipage}
\end{figure}
The second category of tested models comprises BERT \cite{DBLP:journals/corr/abs-1810-04805}, RoBERTa \cite{liu2019robertarobustlyoptimizedbert}, DeBERTa \cite{he2021debertadecodingenhancedbertdisentangled}, and ALBERT \cite{lan2020albertlitebertselfsupervised} models fine-tuned on the dataset with optimized hyperparameters, including a learning rate of 2e-5, maximum sequence length of 128 tokens, batch size of 32, and training with 50 epochs.

The last category explores the capabilities of advanced Large Language Models, including FLAN-T5 \cite{chung2022scalinginstructionfinetunedlanguagemodels}, Llama 3 \cite{dubey2024llama3herdmodels} and  Mistral \cite{jiang2023mistral7b}. For these models, we employed both few-shot learning scenarios (5-shots) and zero-shot classification approaches. We also fine tuned the LLM's using the training data.

We employed an 80-20 split of the dataset, where 80\% of the data was used for training and validation, and the remaining 20\% was used for testing. Within the 80\% training set, we further performed stratified 5-fold cross-validation. The final performance metrics were obtained by averaging the results from all 5 folds.

\section{Results}
Our experimental evaluation demonstrates significant performance differences across model architectures, with transformer-based models and fine-tuned large language models (LLMs) consistently outperforming traditional machine learning approaches. Here, we present a detailed analysis of our findings across different model categories.

\textbf{Traditional Machine Learning Models} \\
As shown in Table~\ref{tab:traditional-ml}, among traditional machine learning approaches, Logistic Regression achieved the best performance with an F1-score of 0.764, followed closely by Random Forest (0.747) and SVM (0.740). Decision Trees showed the lowest performance with an F1-score of 0.671. These baseline results highlight the challenging nature of future reference detection when relying solely on traditional feature engineering approaches.

\textbf{Transformer-Based Models} \\
Transformer-based models demonstrated substantially superior performance compared to traditional approaches. As detailed in Table~\ref{tab:transformer-models}, BERT-base achieved high performance with an F1-score of 0.895. This was followed closely by DeBERTa-base (0.898) and ALBERT-base (0.884). The best performance was achieved by RoBERTa-base, with an accuracy of 0.913 and an F1-score of 0.913. The consistent performance across all metrics 
%(Accuracy, Precision, Recall, and F1-Score) 
suggests robust and balanced prediction capabilities. 

\textbf{Large Language Models} \\
The evaluation of Large Language Models (LLMs) revealed interesting performance patterns in both zero-shot, few-shot, and fine-tuned scenarios, as shown in Table~\ref{tab:llm-models}. Among zero-shot models, Llama 3 achieved the highest accuracy of 0.759, outperforming both FLAN-T5 (0.669) and Mistral (0.740). Few-shot performance generally improved over zero-shot for all models, with Llama 3 again leading with an accuracy of 0.779. 
\begin{table}[h]
\caption{Performance Metrics of Traditional Machine Learning Models}
\label{tab:traditional-ml} 
\centering
\resizebox{0.5\textwidth}{!}{
\begin{tabular}{|c|c|c|c|c|}
\hline
Model & Accuracy & Precision & Recall & F1-Score \\
\hline
Decision Tree & 0.671 & 0.671 & 0.671 & 0.671 \\
Naive Bayes & 0.696 & 0.697 & 0.696 & 0.696 \\
SVM & 0.740 & 0.740 & 0.740 & 0.740 \\
Random Forest & 0.748 & 0.748 & 0.748 & 0.747 \\
Logistic Regression & \textbf{0.764} & \textbf{0.765} & \textbf{0.764} & \textbf{0.764} \\
\hline
\end{tabular}
}
\end{table}

% Table 2: Transformer-Based Models
\begin{table}[h]
\caption{Performance Metrics of Transformer-Based Models}
\label{tab:transformer-models} 
\centering
\resizebox{0.5\textwidth}{!}{
\begin{tabular}{|c|c|c|c|c|}
\hline
Model & Accuracy & Precision & Recall & F1-Score \\
\hline
ALBERT-base & 0.887 & 0.873 & 0.895 & 0.884 \\
BERT-base & 0.892 & 0.903 & 0.887 & 0.895 \\
DeBERTa-base & 0.897 & 0.912 & 0.885 & 0.898 \\
RoBERTa-base & \textbf{0.913} & \textbf{0.921} & \textbf{0.906} & \textbf{0.913} \\
\hline
\end{tabular}
}
\end{table}

% Table 3: Large Language Models
\begin{table}[h]
\caption{Performance Metrics using Large Language Models}
\label{tab:llm-models} 
\centering
\resizebox{0.5\textwidth}{!}{
\begin{tabular}{|c|c|c|c|c|}
\hline
Model & Accuracy & Precision & Recall & F1-Score \\
\hline
FLAN-T5 - Zero-Shot & 0.669 & 0.671 & 0.668 & 0.670 \\
FLAN-T5 - Few-Shot & 0.709 & 0.712 & 0.709 & 0.710 \\
FLAN-T5 - Fine-Tuned & 0.927 & 0.902 & 0.956 & 0.928 \\
\hline
Mistral - Zero-Shot & 0.740 & 0.743 & 0.738 & 0.740 \\
Mistral - Few-Shot & 0.750 & 0.752 & 0.749 & 0.750 \\
Mistral - Fine-Tuned & 0.915 & 0.890 & 0.950 & 0.919 \\
\hline
Llama 3 - Zero-Shot & 0.759 & 0.761 & 0.758 & 0.759 \\
Llama 3 - Few-Shot & 0.779 & 0.781 & 0.778 & 0.779 \\
Llama 3 - Fine-Tuned & \textbf{0.934} & \textbf{0.910} & \textbf{0.960} & \textbf{0.934} \\
\hline
\end{tabular}
}
\end{table}

Fine-tuning these models on our train dataset yielded the most significant improvements in performance. For example, fine-tuned FLAN-T5 achieved an accuracy of 0.927 and an F1-score of 0.928, while fine-tuned Mistral reached 0.915 accuracy and 0.919 F1-score. The highest performance across all LLMs was observed with fine-tuned Llama 3, which achieved an accuracy of \textbf{0.934} and an F1-score of \textbf{0.934}. These results suggest that fine-tuning allows LLMs to significantly enhance their predictive power for future reference detection tasks, yielding even stronger performance than in zero-shot and few-shot setups. The confusion matrices for finetunned Llama 3 and RoBERTa are compared in Fig. \ref{confusion_matrix} and Fig.~\ref{confusion_matrix}, respectively.

\vspace{-.5em}
\section{Conclusion}
\vspace{-.5em}
In this paper we focus on the task of and propose the approach for detecting future references in entity-centric texts, contributing 
%several key advancements 
to the field of temporal information extraction. We curated a new dataset of 19,540 sentences, balanced between future and non-future contexts, centered around high-impact entities from Wikipedia. Our experimental results demonstrate that transformer-based models significantly outperform traditional machine learning methods, with RoBERTa-base achieving the highest performance among the tested transformers with an F1-score of 0.913.
Additionally, the evaluation of Large Language Models in zero-shot, few-shot, and fine-tuned settings revealed the efficacy of fine-tuning. 
%Notably, Llama 3 achieved the best fine-tuned performance among all tested models, with an accuracy and F1-score of 0.934, marking a significant improvement over its zero-shot (0.759) and few-shot (0.779) performance. These findings highlight the potential of LLMs in scenarios where substantial labeled data is available.

%The strong performance of transformer models, even when temporal markers are masked, demonstrates their ability to capture nuanced contextual patterns that indicate future orientation beyond explicit temporal references. The competitive performance of Llama 3 in zero-shot (0.759 F1-score) also points to promising applications in low-resource scenarios where labeled data is limited.

%These findings hold implications for strategic planning, trend analysis, and future-oriented content identification, where automatically detecting future-oriented content about entities can provide valuable insights. 

Future research could explore the integration of domain-specific knowledge, cross-lingual future reference detection, sentence-level bias detection of removal \cite{farber2020multidimensional} and the application of LLM-based approaches to streaming data for real-time entity monitoring. We also plan to provide solutions for extracting sub-components from future-related statements that are useful for aggregating individual forecasts as proposed in \cite{saha2025wisdomcrowdsforecastingforecast}.

%\begin{credits}
%\subsubsection{\ackname} A bold run-in heading in small font size at the end of the paper is used for general acknowledgments, for example: This study was funded by X (grant number Y).

%\subsubsection{\discintname}
%It is now necessary to declare any competing interests or to specifically state that the authors have no competing interests. Please place the statement with a bold run-in heading in small font size beneath the (optional) acknowledgments\footnote{If EquinOCS, our proceedings submission system, is used, then the disclaimer can be provided directly in the system.},for example: 
%The authors have no competing interests to declare that are relevant to the content of this article. 
%Or: Author A has received research grants from Company W. Author B has received a speaker honorarium from Company X and owns stock in Company Y. Author C is a member of committee Z.
%\end{credits}
%
% ---- Bibliography ----
%
% BibTeX users should specify bibliography style 'splncs04'.
% References will then be sorted and formatted in the correct style.
%
%\bibliographystyle{splncs04}
\bibliography{mybibliography}
%

%\bibitem{ref_article1}
%Author, F.: Article title. Journal \textbf{2}(5), 99--110 (2016)

%\bibitem{ref_lncs1}
%Author, F., Author, S.: Title of a proceedings paper. In: Editor,
%F., Editor, S. (eds.) CONFERENCE 2016, LNCS, vol. 9999, pp. 1--13.
%Springer, Heidelberg (2016). \doi{10.10007/1234567890}

%\bibitem{ref_book1}
%Author, F., Author, S., Author, T.: Book title. 2nd edn. Publisher,
%Location (1999)

%\bibitem{ref_proc1}
%Author, A.-B.: Contribution title. In: 9th International Proceedings
%on Proceedings, pp. 1--2. Publisher, Location (2010)
%\bibliographystyle{plain}
%\bibliography{bibliography}

%LNCS Homepage, \url{http://www.springer.com/lncs}, last accessed 2023/10/25

\end{document}